\documentclass[letterpaper, 10 pt, journal, twoside]{IEEEtran}  

\IEEEoverridecommandlockouts                              

\usepackage[usenames,dvipsnames]{color}

\usepackage{multirow}
\usepackage{tabularx}
\usepackage{bm}

\usepackage[pdftex]{graphicx}

\usepackage{balance}
\usepackage{hyperref}
\hypersetup{
  urlcolor=blue}

\usepackage[font=small]{caption}

\IEEEaftertitletext{\vspace{-1.5\baselineskip}}

\usepackage[compact]{titlesec}
\titlespacing{\section}{0pt}{2ex}{1ex}
\titlespacing{\subsection}{0pt}{1ex}{0ex}
\titlespacing{\subsubsection}{0pt}{0.5ex}{0ex}

\begin{document}
\urlstyle{same}
\thispagestyle{empty}
\onecolumn

© 2022 IEEE.  Personal use of this material is permitted.  Permission from IEEE must be obtained for all other uses, in any current or future media, including reprinting/republishing this material for advertising or promotional purposes, creating new collective works, for resale or redistribution to servers or lists, or reuse of any copyrighted component of this work in other works.
\newline

Journal: IEEE Robotics and Automation Letters

DOI: 10.1109/LRA.2022.3147337

URL: \url{https://ieeexplore.ieee.org/document/9699390}
\url{}

\twocolumn

\title{Sim2Air - Synthetic Aerial Dataset for UAV Monitoring}

\author{Antonella Barisic, Frano Petric and Stjepan Bogdan
\thanks{Manuscript received: September 9, 2021; Revised December 7, 2021; Accepted January 11, 2022.}
\thanks{This paper was recommended for publication by Editor Markus Vincze upon evaluation of the Associate Editor and Reviewers' comments.} 
\thanks{Authors are with the University of Zagreb, Faculty of Electrical Engineering  and Computing, LARICS Laboratory for Robotics and Intelligent Control Systems, Unska 3, 10000 Zagreb, Croatia; {\tt\small (antonella.barisic, frano.petric, stjepan.bogdan)@fer.hr}}
\thanks{Digital Object Identifier (DOI): see top of this page.}}

\markboth{IEEE ROBOTICS AND AUTOMATION LETTERS. PREPRINT VERSION. ACCEPTED JANUARY 2022}%
{Barisic \MakeLowercase{\textit{et al.}}: Sim2Air - Synthetic Aerial Dataset for UAV Monitoring}

\maketitle
\setcounter{page}{1}
\begin{abstract}
In this paper we propose a novel approach to generate a synthetic aerial dataset for application in UAV monitoring. We propose to accentuate shape-based object representation by applying texture randomization. A diverse dataset with photorealism in all parameters such as shape, pose, lighting, scale, viewpoint, etc. except for atypical textures is created in a 3D modelling software Blender. Our approach specifically targets two conditions in aerial images where texture of objects is difficult to detect, namely challenging illumination and objects occupying only a small portion of the image. Experimental evaluation of YOLO and Faster R-CNN detectors trained on synthetic data with randomized textures confirmed our approach by increasing the mAP value (17 and 3.7 percentage points for YOLO; 20 and 1.1 percentage points for Faster R-CNN) on two test datasets of real images, both containing UAV-to-UAV images with motion blur. Testing on different domains, we conclude that the more the generalisation ability is put to the test, the more apparent are the advantages of the shape-based representation.

\end{abstract}

\section{Introduction}
\label{sec:introduction}

\IEEEPARstart{S}{ynthetic} datasets offer a potential solution to data-hungry deep learning models that could drive the development of robotics and bring robotics closer to everyday applications. Today, and even more so in the future, robotic systems such as unmanned aerial vehicles (UAVs) often rely on vision-sensors to sense the world and convolutional neural networks (CNNs) to understand the environment and the objects within it. One such task is UAV monitoring, which involves protecting a specific area of interest from unfriendly UAVs. As the number of UAVs in the world increases every year, both commercial and non-commercial, the UAV monitoring systems are becoming an essential part of all high security areas. The performance of such systems depends on the quality and quantity of available data. To address this issue, researchers are exploring the ability of CNNs to learn from synthetically generated data. Unlike real data, which is expensive, time-consuming, and labour-intensive to collect and annotate, synthetic data can be generated automatically and in unlimited quantities. Synthetic annotations are highly accurate and resistant to human error. Moreover, synthetic data enables balanced datasets that cover all desired versions of real data that may be difficult or impossible to obtain. The synthetically generated data may look indistinguishable to humans compared to real data, but this is not the case with CNNs, and this is called the Sim2Real gap. When we add specific imaging conditions of aerial object detection that affect detection accuracy and robustness, we have a Sim2Air gap to bridge.

\begin{figure} [t]
    \centering
    \includegraphics[width=1.0\columnwidth]{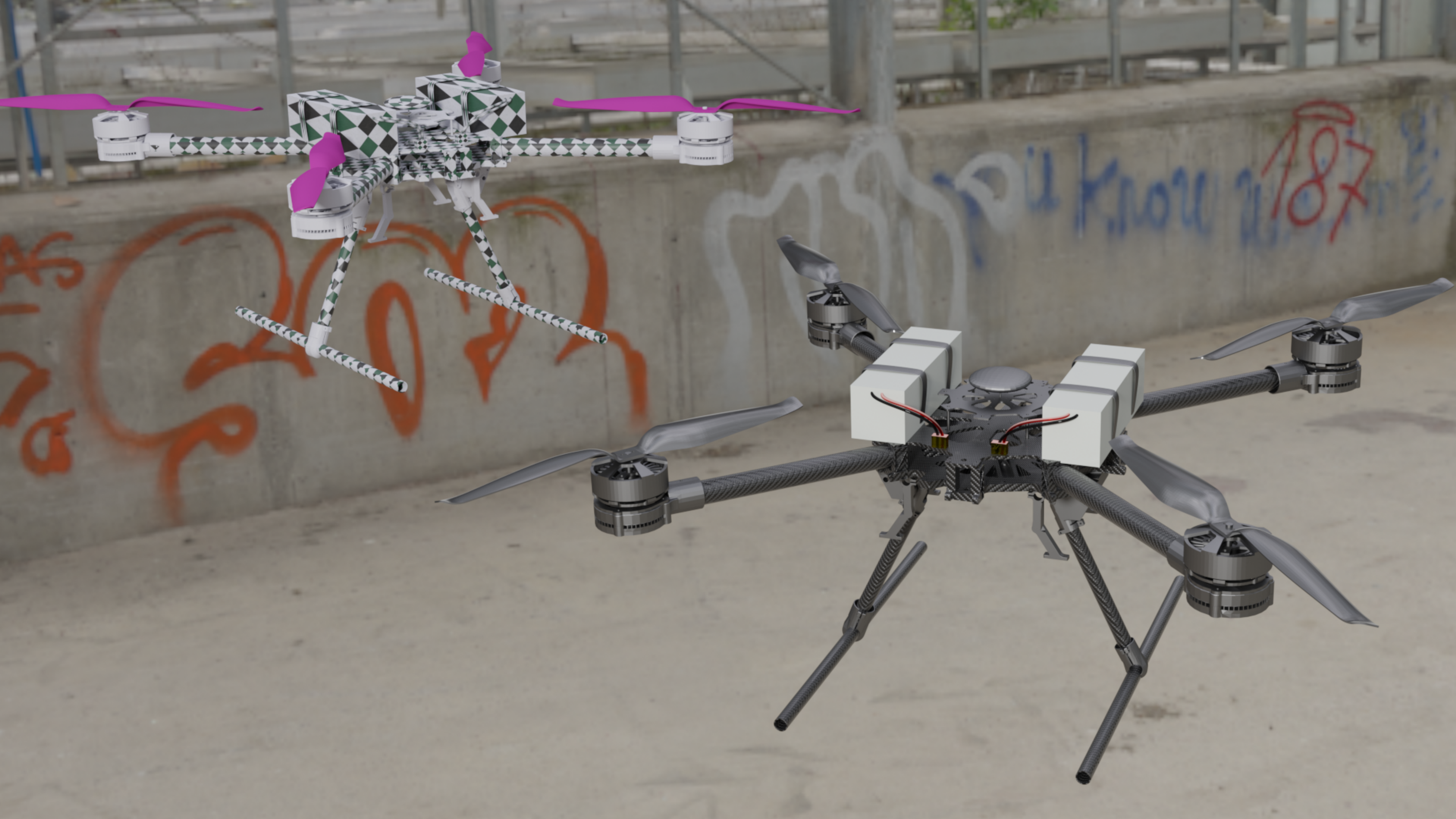}
    \caption{A 3D model of a custom quadcopter Eagle, created in Blender \cite{blender}. The Eagle on the right has realistic textures that match a real aircraft, while the Eagle in the upper left corner has unrealistic textures.}\vspace{-0.5cm}
    \label{fig:eagle}
\end{figure}

The focus of this work is on the creation of synthetic aerial datasets for UAV detection, considering the imaging conditions specific for air-to-air imagery and unconstrained environments. We target two major challenges in aerial object detection, namely long-range detection and detection under changing illumination. The hypothesis of this work is that given the aforementioned challenges, shape-based representation of objects contributes to aerial detection performance. The key contributions of this paper are:
\begin{itemize}
    \item A shape-based representation achieved by randomly assigned unrealistic textures to improve the performance of aerial object detection in the face of changing illumination conditions in unconstrained environments and the difficult detection of objects that occupy only a small portion of the image, which is very common in aerial object detection;
    \item A procedural pipeline for generating synthetic aerial dataset for UAV monitoring with diversity of models, backgrounds, lightning, times of day, weather conditions, positions and orientations, camera pan and tilt angles, and distances from camera to object;
    \item A first publicly available synthetic dataset for object detection of UAVs.
\end{itemize}

The remainder of the paper is organized as follows: Section \ref{sec:related_work} presents recent state-of-art works and findings, while Section \ref{sec:aerial_od} gives a problem description of aerial object detection. Section \ref{sec:method} describes the proposed method to bridge the Sim2Air gap, while experimental validation of the proposed hypothesis is described in Section \ref{sec:results}.
\section{Related work}
\label{sec:related_work}

\textbf{Synthetic datasets.} Synthetically generated data has the potential to completely overcome the problem of tedious manual creation of large annotated datasets for training data-driven deep learning models. Currently, there are three popular paradigms in the synthetic dataset community: combining real and synthetic data outperforms pure real data, domain adaptation, and domain randomization. The first one has been validated by many researchers \cite{Nowruzi2019HowMR, Linder2020,Tremblay2018}, while the other two paradigms continue to be the subject of research. Domain adaptation techniques aim to bridge the Sim2Real gap by minimising the difference between synthetic and real data, while domain randomization techniques aim to randomise synthetic data to the point where the real world is considered just another synthetically generated variant. Tobin et al. \cite{Tobin2017} proposed domain randomization in task of object grasping by randomizing the RGB values of the object texture. A year later, Trembley et al. \cite{Tremblay2018} proposed randomizing lighting, pose, and textures in an unrealistic way and adding flying distractors. Both works show compelling performance compared to CNNs trained on purely real data. Another interesting direction proposed in \cite{Geirhos2019ImageNettrainedCA} is to explore the texture bias of neural networks. They have shown that texture plays an important role, even more than global object shapes, and that shape-based representations might be more informative. With respect to implementation, there are multiple approaches to generating synthetic data, which can be summarized in two broad categories: 1) deep learning: using models such as Generative Adversarial Networks and Variational Autoencoders \cite{gan_vae} to generate synthetic data; 2) 3D rendering: using 3D modelling software such as Unity, Unreal or Blender to generate synthetic data. Two most common approaches in synthetic data generation using 3D rendering are aimed towards using realistic models and/or realistic sensors \cite{synthetic_overview}. Herein, we propose a procedural pipeline that uses realistic models of the object, augmented with unrealistic textures to emphasize geometric shape of the object in the process of learning.

\textbf{Visual UAV monitoring.} Monitoring UAV activity in a predefined area of special interest requires three technologies: detection, interdiction, and evidence collection \cite{Taha2019}. The use of image sensors for UAV surveillance, especially for the detection part, has attracted much attention recently. One approach to general UAV-to-UAV detection is to collect and annotate a large amount of real images using different UAV models and types, at different times of the day and under different weather conditions, in different locations to cover a variety of backgrounds, in different positions and orientations, and at different scales and viewpoints. There are several works \cite{Vrba2020, Barisic2019, Yavariabdi2021, Wyder2019} that use the above approach to train the detector on real images that had to be collected and manually labelled over a long period of time. In the absence of publicly available datasets, few works \cite{Chen2017, Aker2017} have trained UAV detectors on artificial datasets created by pasting object models on background images. Although this solves the data problem, it does not address the specific imaging conditions for drones, does not provide a balanced dataset covering all possible variations of the real world, and suffers from unrealistic lighting due to the simple insertion technique. Therefore, 3D modelling software such as Blender is a more resourceful solution to create synthetic datasets in a controlled manner. Controlling the diversity in a dataset using a procedural pipeline for rendering, like in \cite{denninger2019blenderproc}, allows to cover all real-world variations and create balanced datasets. Most similar to our approach is the work of Peng et al. \cite{Peng2018} who propose a procedural pipeline using physically-based rendering toolkit (PBRT) to render as photorealistic images as possible. However, our approach differs not only in having a larger number of models, a different rendering engine and a lightweight detector, but also in emphasising shape-based representation to reduce texture bias and improve the accuracy and robustness of aerial object detectors.
\section{Aerial object detection}
\label{sec:aerial_od}

Object detection from images captured on-board UAV presents additional challenges due to the nature of the aircraft's motion and operating conditions. Aerial perspective means that the detector must be able to detect objects at different pan and tilt angles, at different altitudes, and in 360 degrees. Therefore, it is necessary to distinguish between detection in still images captured by humans, also called generic object detection, and detection on aerial images captured from a moving aircraft. From a robotics perspective, object detection of specific instances such as other UAVs, plants, buildings, etc. is of greater interest. 

\begin{figure} [htb]
    \vspace{-0.2cm}
    \centering
    \includegraphics[width=1.0\columnwidth]{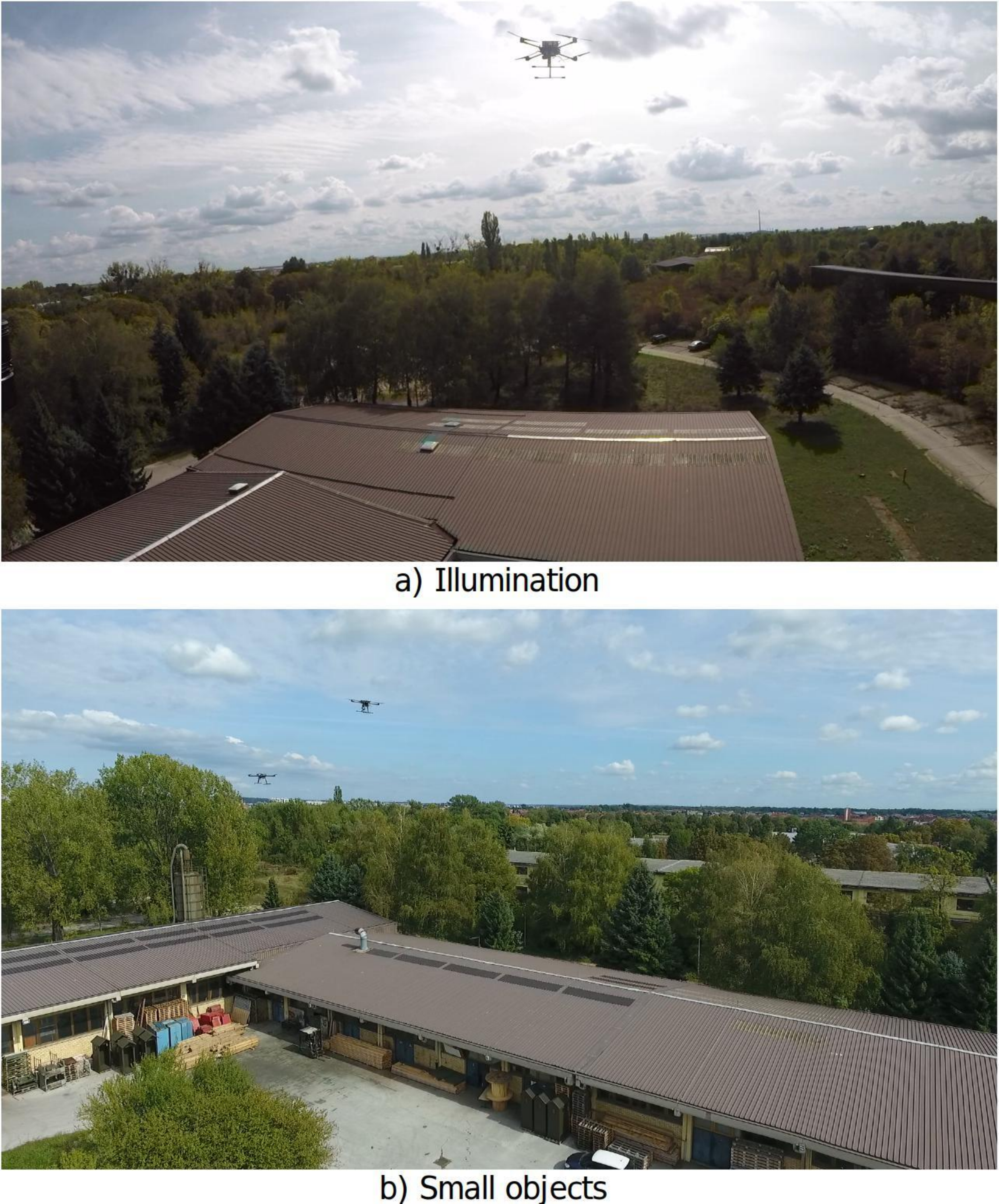}
    \caption{Imaging conditions of aerial object detection from UAV-Eagle dataset.}\vspace{-0.25cm}
    \label{fig:imaging}
\end{figure}

Intra-class variations, which are the main factor affecting accuracy and challenging the generalisation ability of the detector, can be divided into two categories \cite{Liu2019}: intrinsic factors and imaging conditions. Intrinsic factors are variations in appearance within a class, such as different sizes, shapes, textures, etc., which can be addressed by synthetic datasets with high diversity. On the other hand, variations in appearance caused by imaging conditions such as lighting, pose, weather conditions, background clutter, and viewing angles are an innate consequence of unstructured environments. Since UAVs typically operate in such unstructured dynamic environments, aerial imaging conditions must be considered when preparing synthetic datasets. Two problems that are particularly specific to aerial object detection, namely long-range detection and detection under changing illumination, are shown in Fig. \ref{fig:imaging} using a UAV-Eagle dataset \cite{barisic2021brain} acquired in an unstructured environment. Under varying illumination the same object can appear very different \cite{Adini1997}, especially in color, depending on the relative position and orientation between the object and the light source in the scene. On the other hand, objects located at a great distance occupy only a small part of the image, which makes them more difficult to detect because they contain less information about their appearance and require more precise localization. In both problems, shape is more visible and prominent than texture. Therefore, our goal is to guide the object detector towards shape-based detection in order to bridge the Sim2Air gap.

\section{Methodology}
\label{sec:method}
In this section, the proposed methodology is described in detail and the main components are outlined.

\subsection{General pipeline}

All synthetic datasets in this paper were created in Blender, an open source software for 3D modelling. Blender offers a wide range of tools for modelling all kinds of objects, surfaces, lighting, environments, etc. There are two options available as rendering engines: Eevee and Cycles. The first is for real-time rendering, the second for ultra-realistic ray-trace rendering. In our pipeline, we used Cycles for rendering with 512 samples of path-tracing the light for each pixel. An important feature of Blender is its built-in Python interpreter. User can access Blender objects and tools via Python scripts and manipulate them in a variety of ways. This allows us to take full advantage of the automation and scalability properties of synthetic datasets. For example, our pipeline allows the user to choose how many textures or how many random panning angles to use when rendering a synthetic dataset. Other parameters are scaled accordingly to render the desired number of images. Therefore, scalability of the dataset is easily feasible with the proposed procedural pipeline. In parallel with image rendering, all objects in the current image are automatically and precisely labelled with bounding boxes enclosing each object. The labels are stored in a appropriate format for later training of the detector.

Considering all the aspects of aerial object detection mentioned in Section \ref{sec:aerial_od}, and to account for the diversity of the real world, we implement variations in the following components within the procedural pipeline:
\begin{itemize}
    \item 3D models of different quadcopters and hexacopters,
    \item number of objects in a single image,
    \item environment maps with different lighting conditions such as daylight, partly cloudy, twilight, etc.
    \item distance between camera and objects, i.e. scale of objects,
    \item angle of the camera (pan, tilt and yaw) with respect to the world origin,
    \item location of the camera with respect to the world origin,
    \item a mixture of atypical textures.
\end{itemize}

\begin{table}[htb]
\caption{Details of the datasets used for training: \textbf{S}ynthetic \textbf{Eagle} \textbf{B}aseline (S-Eagle-B), \textbf{S}ynthetic \textbf{Eagle} with \textbf{T}extures (S-Eagle-T), \textbf{S}ynthetic \textbf{UAV}s with \textbf{T}extures (S-UAV-T) and \textbf{R}eal \textbf{UAV} (R-UAV); and for testing: UAV-Eagle, T$_1$ and T$_2$.}
\label{tab:datasets}
\centering
\begin{tabular}{l|cccc } \hline

\multicolumn{5}{c}{Train datasets} \\ \hline \hline

& \textbf{S-Eagle-B}
& \textbf{S-Eagle-T}
& \textbf{S-UAV-T}
& \textbf{R-UAV}  \\ \hline

Models & 1 & 1 & 10 & $\sim$20\\ 
Backgrounds & 10 & 10 & 10 & $\infty$ \\ 
Textures & 1 & 32 & 32 & N/A\\ 
Pitch $[^\circ]$&  [-45, 45] &  [-45, 45] &  [-45, 45] & N/A \\
Roll $[^\circ]$&  [-45, 45] &  [-45, 45] &  [-45, 45] & N/A \\ 
Yaw $[^\circ]$ & [0, 360] & [0, 360] & [0, 360] & N/A\\ 
Distance $[m]$ & [2, 20] & [2, 20] & [2, 20] & N/A \\ 
Image No.& 32 000 & 32 000 & 52 500 & 11 700\\ \hline \noalign{\vskip 1mm} 
\multicolumn{5}{c}{Test datasets} \\ \hline \hline

\end{tabular}

\begin{tabularx}{0.92\columnwidth}{Xc|cccX}

& 
& \textbf{UAV-Eagle}
& \textbf{T$_1$}
& \hspace{0.35cm} \textbf{T$_2$} & \\ \hline 

& Models & 1 & $\sim$20 & \hspace{0.35cm} 1 & \\ 
& Backgrounds & 3 & $\infty$ & \hspace{0.35cm} 5  &\\ 
& Image No. & 510 & 1300 & \hspace{0.35cm} 285 &\\ \hline
\end{tabularx}

\centering
\end{table}

Each model is rendered in all environment maps from the set, and each texture from a set of mixed textures is assigned to the model. Then for each iteration, a predefined number of random values for pan, tilt, and the distance between the camera and the objects are selected. With such configuration in a scene, a set of images is rendered by animating the yaw angle of the camera around objects. All components except texture are modelled to reflect the real world as closely as possible. Details of the parameters of each dataset used to train the aerial object detector are given in the Table \ref{tab:datasets}, where the first letter of the dataset name indicates the type of data (synthetic or real), the middle part indicates the type of models used (Eagle quadcopter only or multiple UAV models), and the last letter indicates the type of technique, if any, used to create the dataset (e.g., -T for texture randomization).

\subsection{Models}

The first model in this paper is the Eagle quadcopter. The Eagle is a custom aerial platform that is used for a variety of tasks as its modular design allows for different sensor configurations. A custom UAV presents a more difficult problem than other commercial UAVs because data (images) collection is more challenging. One should collect a large number of air-to-air images to realistically reflect the imaging conditions of UAV detection from the air. To overcome this problem, we propose to use 3D models of custom aerial platforms such as Eagle. The photorealistic model of Eagle shown in Fig. \ref{fig:eagle}, which closely resembles an actual aircraft, was created using standard modelling techniques in Blender.

\begin{figure} [htb]
    \centering
    \includegraphics[width=1.0\columnwidth]{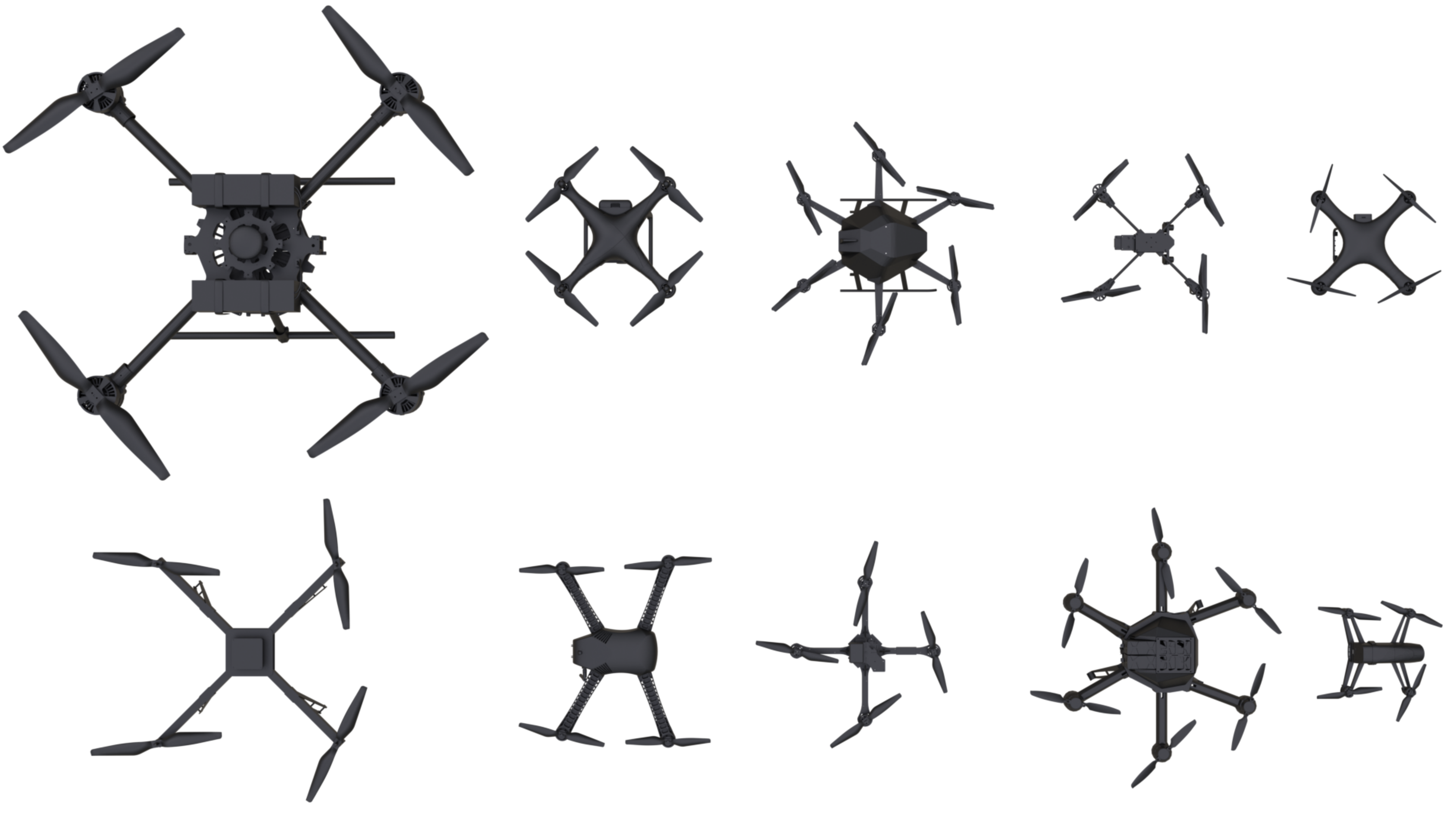}
    \caption{All 3D models of UAVs used to create a synthetic dataset for aerial object detection.}\vspace{-0.25cm}
    \label{fig:models}
\end{figure}

In addition to the 3D model of Eagle, another 9 models of quadcopters and hexacopters were used to create a synthetic aerial dataset for UAV monitoring. Three models were found online \cite{cgtrader}, while the other six models were obtained from Gazebo simulations, mainly from the RotorS \cite{Furrer2016} library. All these together result in a set of 10 different UAV models, which are shown in Fig. \ref{fig:models}.

\subsection{Textures}

\begin{figure} [t!]
    \centering
    \includegraphics[width=1.0\columnwidth]{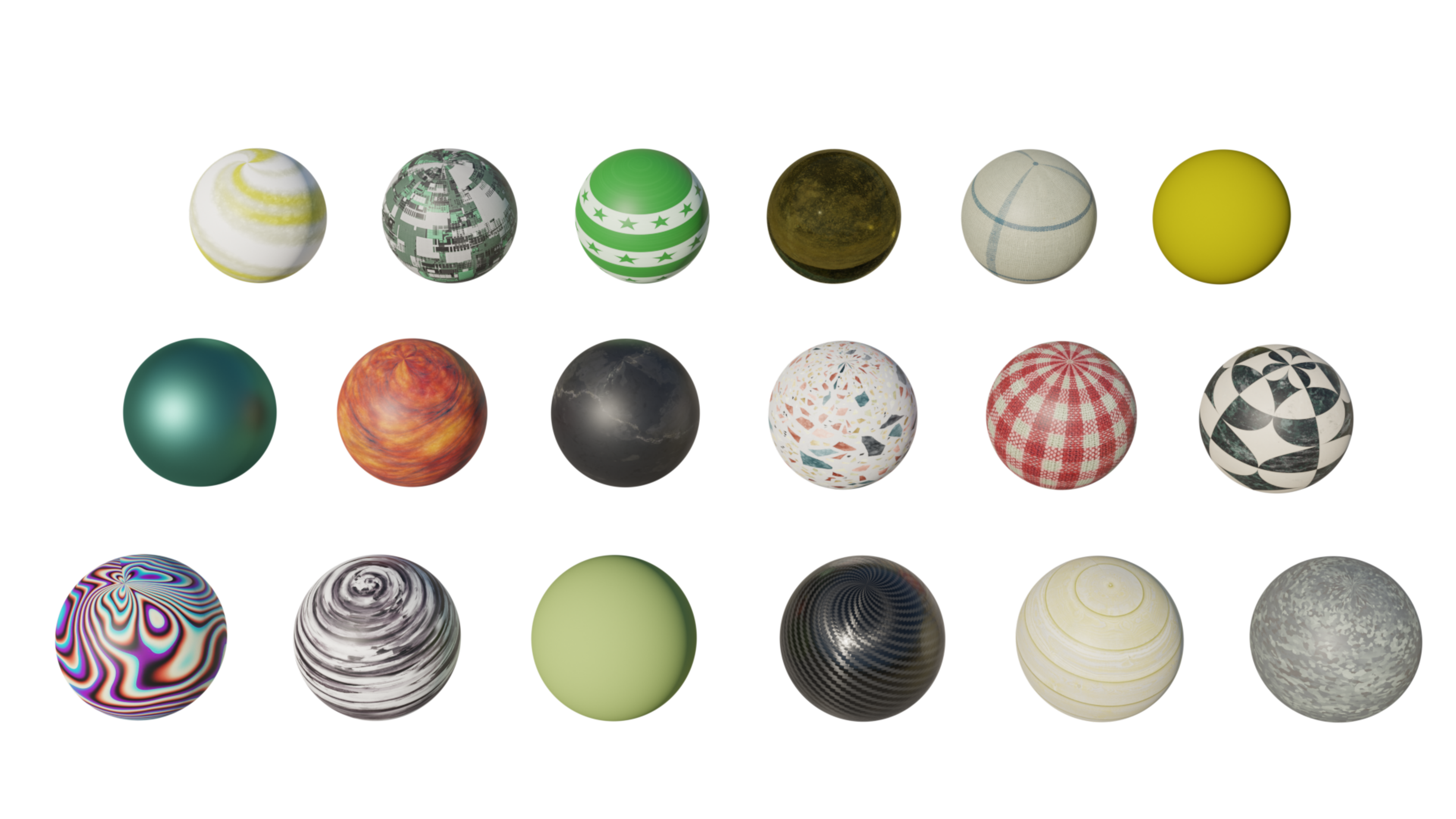}
    \caption{Sample of textures used for a texture-randomization technique.}\vspace{-0.25cm}
    \label{fig:tex}
\end{figure}

Each time the generation of a synthetic dataset is started, a different mixture of textures is created (see Fig. \ref{fig:tex}), which can later be assigned to objects in a scene. Some of the textures are procedurally created by an algorithm that combines different material properties with random colors. Different material properties are implemented by so-called shaders, which determine how light is scattered across the surface of the object. For a predefined number of procedurally generated textures, a shader is selected from four shaders, namely diffuse, glossy, glass and translucent bidirectional scattering distribution function (BSDF). Each shader implements a different mathematical function that describes the reflection, refraction, and absorption of light. A diffuse shader is the most realistic material for UAVs, while the others contribute to the unrealism of the textures. To demonstrate the irrelevance of colors to the detector, each material is assigned a random RGB value for color. The second part of the texture mixture is created by importing image files with textures that are atypical for UAVs, e.g. bathroom tiles, lollipop texture, Christmas ornaments, kitchen towel, wood, chip texture, marble, metal, etc. In the end, a realistic texture of carbon-made UAV is added to the mixture in order to treat the real texture as another random variation. The thus created mixture of textures using both monochromatic and patterned textures, typical and atypical colors, artificial and man-made textures, reflective and non-reflective materials, etc. leads the detector to pay more attention to the shape of the UAV.

\subsection{Environment maps}

Instead of modelling a series of complex scenes, a set of environment maps is used to replicate typical unstructured environments of UAVs. The environment maps are implemented using High Dynamic Range Imaging (HDRI), a 360-degree panoramic image with extensive brightness data. The HDRI maps not only serve as a background for the scene, but also provide realistic illumination of the scene. This is especially useful in the case of custom aerial platforms, as images containing them are difficult to collect. Collecting images could take months to cover different weather conditions such as sunny, cloudy, foggy, etc., take place at different times of the day to cover all lighting conditions, and would also require at least two UAV operators. All of these problems are addressed with a set of 10 HDRI maps, acquired from Polyhaven \cite{polyhaven}, with different lighting conditions and different types of outdoor environments.

\section{Results}
\label{sec:results}

To verify our claims, a general object detector suitable for on-board UAV processing was trained on several datasets. The experiments are divided into two parts. First, we train the detector on datasets based on a single custom model of UAV, with and without applying random unrealistic textures, to show the impact of the proposed technique on the generalisation and robustness of the aerial object detector. Second, we create a larger dataset with different UAV models and compare the performance between synthetic and real data in the UAV monitoring task. The quantitative and qualitative results of how our method improves the detection of objects in the air are presented below. The synthetic datasets are available at \href{https://github.com/larics/synthetic-UAV}{\color{blue}https://github.com/larics/synthetic-UAV} and the video of our approach can be found at \href{https://www.youtube.com/watch?v=7pPGEk8t_Tw}{\color{blue}https://www.youtube.com/watch?v=7pPGEk8t\_Tw}.

\begin{figure*} [htb]
    \centering
    \includegraphics[width=1.0\textwidth]{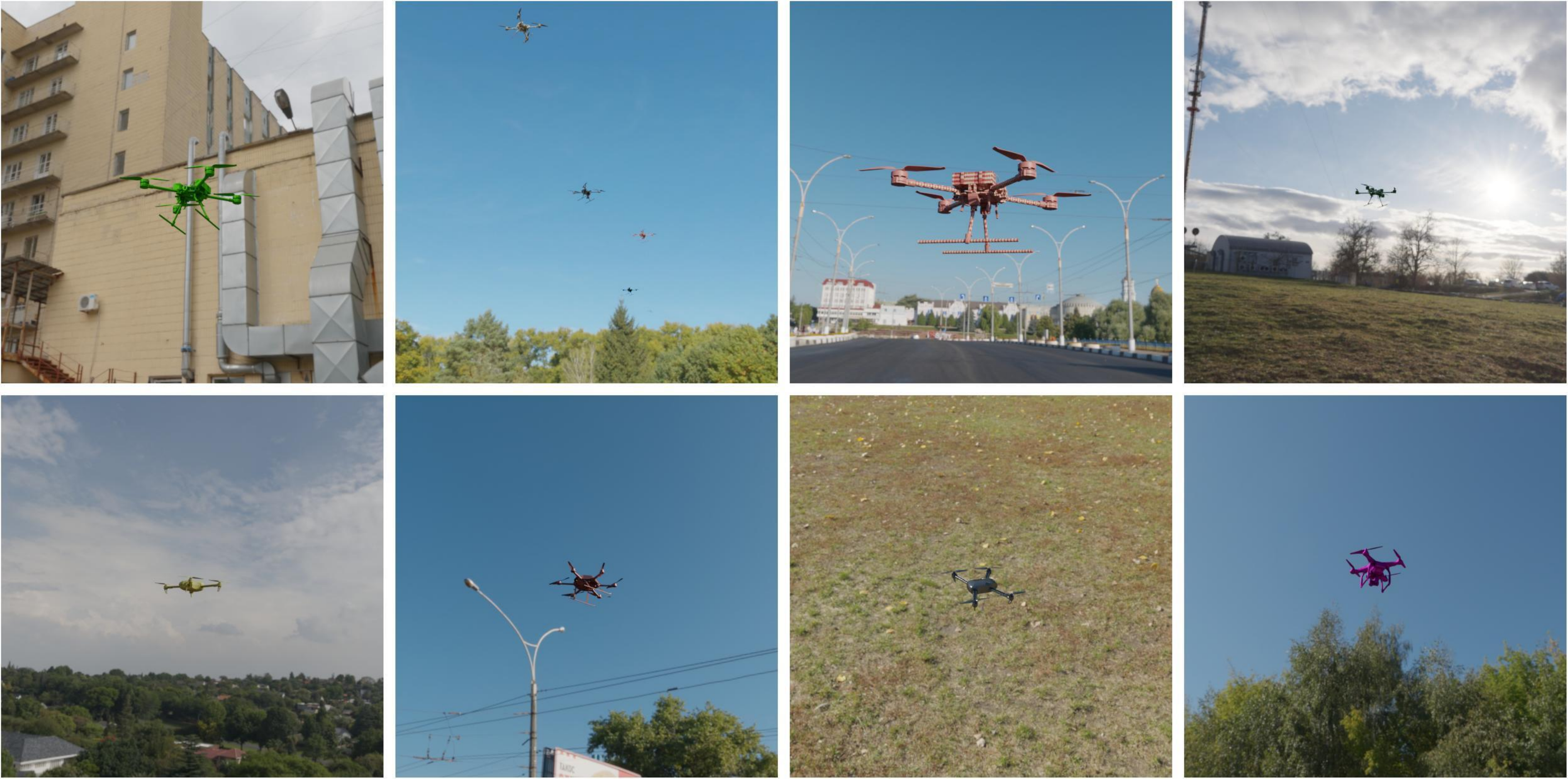}
    \caption{Examples of images in the synthetically generated datasets S-Eagle-T (top row) and S-UAV-T (bottom row). Both datasets contain randomized textures, while S-UAV-T contains 9 other UAV models in addition to Eagle quadcopter.}\vspace{-0.25cm}
    \label{fig:samples}
\end{figure*}

\subsection{Training of object detector}
\label{subsec:yolo}

In this paper, two object detectors, each representative of its category, are trained and evaluated for the task of aerial object detection. The first one comes from a well-established family of one-stage detectors called You Only Look Once (YOLO) \cite{Redmon2016, bochkovskiy2020yolov4}. YOLO detectors treat object detection as a regression problem and use features from the entire image to detect objects, implicitly including contextual information. There are two advantages of YOLO that led us to choose it as an object detector. The first important advantage is the low inference time, which accommodates the limited computational resources of on-board processing. The second advantage is the accessibility of the network implementation along with the continuous development and wide community support, which allows easy portability to any application. In selecting the detector, we looked for a lightweight network and therefore selected a lightweight version, called Tiny, of the fourth generation of YOLO networks. Taking into account that bird's eye objects usually occupy only a small part of the image, the network architecture was slightly modified by adding another YOLO layer to better detect both small and large objects. On the other hand, we chose the Faster R-CNN \cite{Ren2017}, a representative of the two-stage object detectors, as the second detector to show that the proposed method does not depend on the type of detector. We utilize a base model with a backbone combination of ResNet-50 and Feature Pyramid Network with a $3x$ learning rate scheduler. In general, the Faster R-CNN is a more accurate detector than YOLO, but it requires significantly more time for inference and is therefore generally not suitable for on-board processing.

All training and dataset rendering was performed on a computer equipped with a Intel Core i7-10700 CPU @ 2.9 GHz x 16, an Nvidia GeForce RTX 3090 24 GB GPU, and 64 GB RAM. The YOLO detector was trained within Darknet framework and Faster R-CNN within Detectron2 \cite{wu2019detectron2} framework, both using default anchors and starting from weights pretrained on the COCO dataset \cite{Lin2014}. For training, we used an image size of 608, a learning rate of 0.00261, a momentum of 0.9 and a decay of 0.0005. Each model was trained for 200 epochs, using a batch size of 64 for YOLO and a batch size of 16 for Faster R-CNN. The best weights were determined using an early stopping method based on the mean average precision (mAP) to avoid overfitting. The training parameters were the same for all datasets.

\subsection{Texture-invariant dataset of Eagle UAV}

To test our hypothesis that a shape-based, i.e. texture-invariant, representation of UAVs boosts performance of aerial object detectors, we created two datasets: \textbf{S}ynthetic \textbf{Eagle} \textbf{B}aseline (S-Eagle-B) and \textbf{S}ynthetic \textbf{Eagle} with \textbf{T}extures (S-Eagle-T) dataset. Both datasets were created using a single UAV model, the custom aerial platform Eagle, and contain the same number of images. The main difference is that the baseline dataset was created with a photorealistic texture that mimics the real aircraft: a carbon frame with a synthetic plastic texture for smaller parts of the model, as shown in the lower right corner in Fig. \ref{fig:eagle}. The S-Eagle-T dataset was created with 32 unrealistic textures to accentuate the geometric shape of the UAV. Samples from the S-Eagle-T dataset are shown in the top row of Fig. \ref{fig:samples}.

\begin{figure} [htb]
    \centering
    \includegraphics[width=1.0\columnwidth]{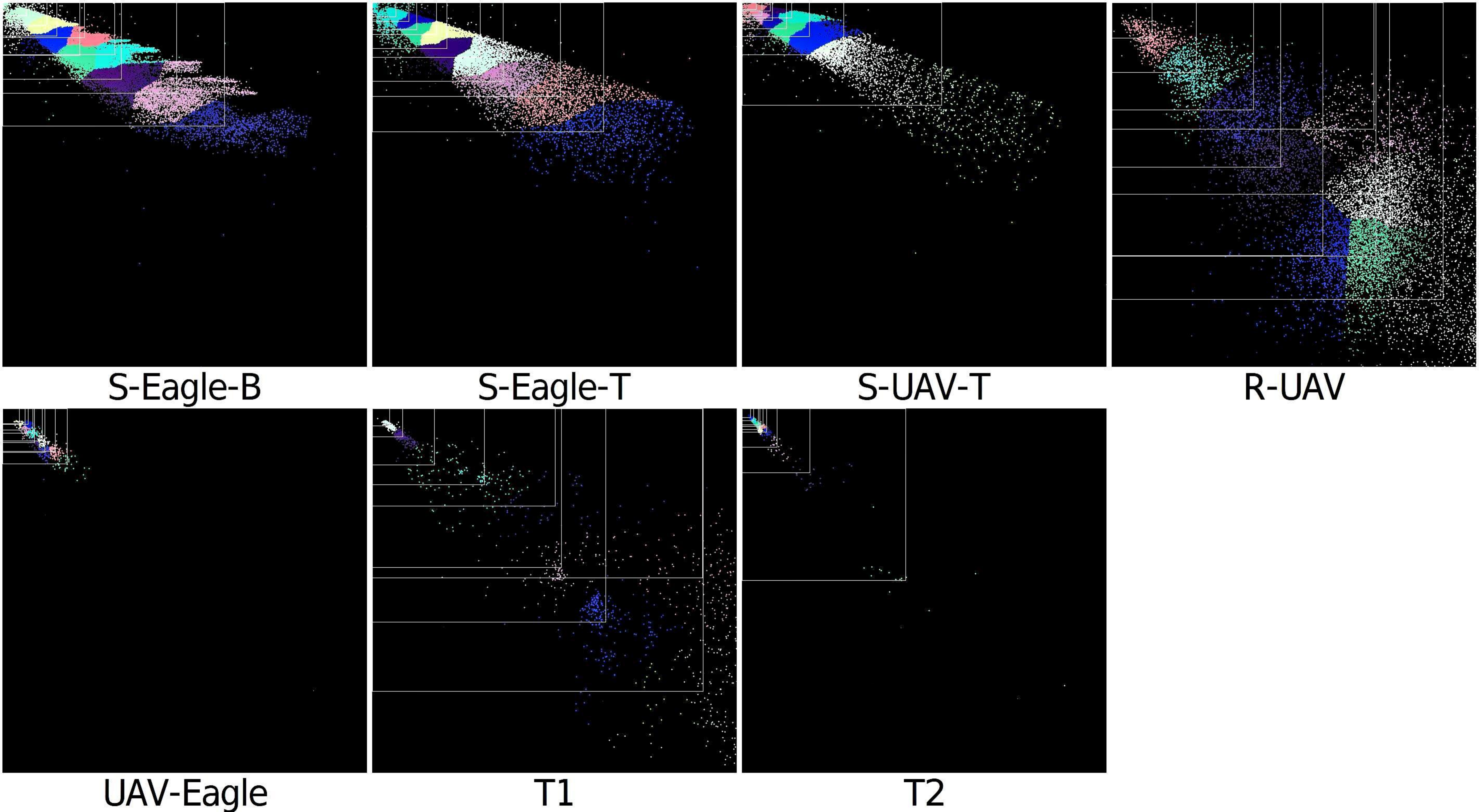}
    \caption{Distribution of data in the training and testing datasets obtained by the k-means++ algorithm. The colors indicate the 9 clusters of computed anchor boxes that best correspond to the actual size of the objects in the dataset. The coordinates of each pixel correspond to the width and height of each bounding box.}\vspace{-0.15cm}
    \label{fig:cluster}
\end{figure}

The detectors trained on S-Eagle-B and S-Eagle-T datasets were evaluated on three different test datasets, all consisting of real images only. The details are given in Table \ref{tab:datasets}. The first dataset is UAV-Eagle \cite{barisic2021brain}, which contains images of real Eagle UAV in an unconstrained environment with pronounced illumination effects, distant objects, and a highly cluttered background. The second test dataset is part of our previous work \cite{Barisic2019, barisic2021brain}, where we collected a large number of real images from the Internet. The labelling was done partly by manual labelling and partly by pseudo-labelling. This dataset consists of 13 000 images which are divided into two parts: a training dataset called R-UAV containing 90\% of the images, and a test dataset T$_1$ containing the remaining 10\%. Some of the images in T$_1$ were taken in the studio, some indoors, some outdoors, and most of the images feature commercial drones. The third and last dataset for evaluation, called T$_2$, is a test dataset from \cite{Vrba2020}. The T$_2$ dataset consists of real images of the author's custom UAV taken in different unstructured environments. The datasets have different domains, from the style of the images to the type of objects and the size of the objects in the images. The T$_2$ dataset is the most difficult in terms of detecting small objects, as can be seen in Fig. \ref{fig:cluster}, followed by UAV-Eagle. An additional challenge in the UAV-Eagle and T$_2$ datasets are images of UAVs captured from UAVs, which introduce significant amount of motion blur. The T$_1$ and R-UAV datasets, on the other hand, contain larger objects.

\begin{table}[htb]
\caption{Evaluation results of detectors YOLOv4 Tiny and Faster R-CNN trained on baseline and proposed approach.}
\label{tab:map}
\begin{center}
\begin{tabular}{l|c|c|c|c}
\hline
Detector
& Train dataset
& UAV-Eagle
& T$_1$
& T$_2$  \\ \hline \hline

\multirow{2}{*}{YOLO} 
& S-Eagle-B & \textbf{86.84\%} & 30.11\% & 64.76\%  \\
& S-Eagle-T & 85.65\% &   \textbf{47.16\%}   & \textbf{68.47\%} \\ \hline

\multirow{2}{*}{Faster R-CNN} 
& S-Eagle-B & \textbf{92.28\%} & 65.50\% & 64.68\%   \\ 
& S-Eagle-T & 89.01\% &   \textbf{85.39\%}   & \textbf{65.75\%} \\ \hline

\end{tabular}
\end{center}
\end{table}

\begin{figure} [htb]
    \centering
    \includegraphics[width=1.0\columnwidth]{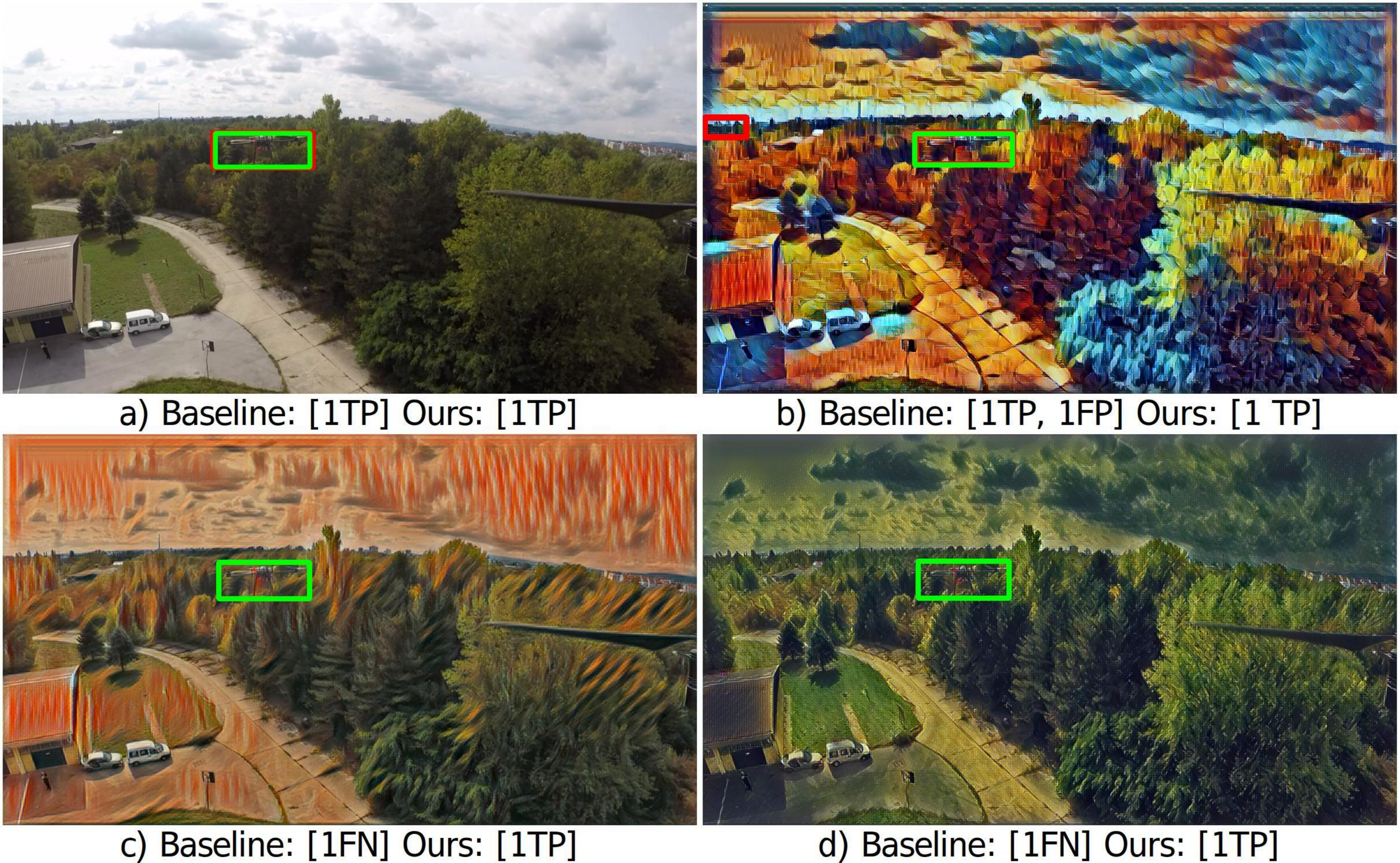}
    \caption{Detection results of the baseline S-Eagle-B (red) and our method S-Eagle-T (green) on images with different textures. While both detectors accurately detect the original image (a), the proposed method performs better than the baseline method on stylized images (b,c,d) without texture cues. TP stands for true positive, FP for false positive and FN for false negative detection.}\vspace{-0.25cm}
    \label{fig:NST}
\end{figure}

The evaluation results in terms of the mean Average Precision (mAP) with an Intersection over Union (IoU) threshold of 0.5 are presented in Table \ref{tab:map}. On the UAV-Eagle dataset, the detectors trained with realistic texture achieve a better mAP value than the detector trained with random atypical textures. This is expected since the synthetic model and texture fit the test dataset almost perfectly. In the test dataset T$_1$, however, the detectors trained with a shape-based representation outperform the baseline by a large margin, by 17 percentage points in mAP for YOLO and 20 percentage points for Faster R-CNN. The significant difference indicates that by guiding detector towards shape-based representation, we have improved the performance of the aerial object detector, more specifically its generalisation ability and robustness. The T$_1$ dataset is very different in style from the synthetic datasets on which the detectors were trained. For this reason, the baseline detector with only one texture experiences a large texture-bias that degrades its generalisation ability, while the S-Eagle-T detector shows improved accuracy on previously unseen UAVs. To confirm that the detector trained on S-Eagle-T gives more importance to the shape of the UAV, e.g., the frame, propellers, and landing gear, we perform additional tests with YOLO detector. We eliminate texture cues while preserving the shape of the object from the image using a neural style transfer algorithm \cite{tensorfire}. While the S-Eagle-B detector struggles without texture cues, the qualitative results in Fig. \ref{fig:NST} show that the S-Eagle-T detector handles texture variations well.

In the third evaluation, the S-Eagle-B and S-Eagle-T detectors are compared on the test dataset T$_2$, which contains images of a custom UAV that were not in the training data of any detector. Using a texture-invariant approach, an increase in mAP of 3.71 percentage points for YOLO and 1.07 for Faster R-CNN is observed. This is further confirmation that our approach improves the performance of the aerial object detector, as it has better generalisation ability and robustness to small objects on T$_2$ dataset. The larger increase in mAP when evaluating on the T$_1$ dataset can be explained by the larger difference in domain between the T$_1$ and synthetic datasets. Thus, the more the ability to generalise is put to the test, the more evident becomes the influence of shape-based representation on the improvement of accuracy.

\subsection{Synthetic vs. Real}

For application in UAV monitoring, we have created a synthetic dataset of 52 500 images with 10 different UAV models using the proposed procedural pipeline and texture randomization technique. We named this dataset \textbf{S}ynthetic \textbf{UAV}s with \textbf{T}extures (S-UAV-T). Sample images are shown in the bottom row of Fig. \ref{fig:samples}. For the following experiments, the YOLO detector is utilized since it is suitable for on-board UAV processing. Following the procedure described in subsection \ref{subsec:yolo}, the detector is trained on S-UAV-T synthetic dataset and the results are presented in Table \ref{tab:map2}. Although we expected to obtain better results with more models, the evaluation on test datasets with real images showed the opposite. Overall, the detector trained on S-Eagle-T outperforms the one trained on S-UAV-T, suggesting that a carefully selected model is sufficient. Other factors that could account for this result are the higher prevalence of small objects in the S-UAV-T dataset and overfitting to synthetic data.

\begin{table}[htb]
\vspace{0.1cm}
\caption{Evaluation results of the YOLO detector trained on synthetic data, trained on real data, and trained on synthetic data and then fine-tuned on real data.
} 
\label{tab:map2}
\begin{center}
\begin{tabular}{l|c|c|c }
\hline 
Train dataset
& UAV-Eagle
& T$_1$
& T$_2$  \\ \hline \hline

S-UAV-T & 73.80\%  & 42.41\%   & 68.83\%  \\ 
S-UAV-T$_f$ & 95.24\%   &  \textbf{93.86} \% &  71.54\% \\
S-Eagle-T$_f$ & \textbf{95.90\%}  &  91.23\% & \textbf{76.44\%} \\
R-UAV & 90.38\%  & 91.31\% & 65.73\% \\ \hline

\end{tabular}
\end{center}
\vspace{-0.25cm}
\end{table}

Just as we addressed the gap between aerial and generic object detection in the previous section, we now apply the fine-tuning technique to close the gap between synthetic and real world. The detectors trained on the S-Eagle-T and S-UAV-T datasets are fine-tuned for 20 epochs on real images from the R-UAV dataset with decreased learning rate of 0.00161. To allow a full comparison of the performance of the proposed approach, YOLO detector was also trained on the R-UAV dataset using only real images. The detection results of the trained detectors are shown in Table \ref{tab:map2}. The fine-tuned datasets, indexed by $f$, show comparable performance. The S-Eagle-T$_f$ detector performs better on datasets whose domain is custom-made quadcopters, while S-UAV-T$_f$ detector performs better on T$_1$ dataset that contains more UAV models. Moreover, the results show that both detectors trained on a balanced synthetic datasets, built specifically to include diverse poses of the object under different lighting and backdrop, and fine-tuned on real data, perform better or equally well as the detector trained purely on a collection of images obtained and annotated through tedious work. 

Apart from the fact that the combination of synthetic and real data is better for all three test datasets, the biggest difference is seen in the T$_2$ dataset, where S-Eagle-T$_f$ outperforms the R-UAV detector by almost 11 percentage points. We conclude that there are two reasons for this. First, as seen in Fig. \ref{fig:cluster}, the T$_2$ dataset is most difficult in terms of small object detection which is why our texture-invariant detector performs drastically better. Second, just like UAV-Eagle, the T$_2$ dataset contains only one type of custom UAV, which are not included in the R-UAV dataset and differ in appearance from the commercial and popular UAVs that make up the bulk of the R-UAV dataset. Therefore, the generalisation capability of detector trained on R-UAV is limited. We strongly believe that these results confirm that introduction of texture-randomization through synthetic datasets places higher importance on the shape of the UAV, which in turn yields improved detection results.

\subsection{Discussion}

\begin{table}[htb]
\caption{Robustness against challenging illumination conditions. Comparison of the baseline and the proposed method on modified T2 and UAV-Eagle test datasets.
} 
\label{tab:illum}
\begin{center}
\begin{tabular}{l|l|c|c }
\hline 
Test dataset
& Method
& S-Eagle-B
& S-Eagle-T \\ \hline \hline

\multirow{2}{*}{UAV-Eagle}  & Original & \textbf{86.84} & 85.65 \\
 & Challenging Illumination & 72.35 & \textbf{76.95} \\ \hline
 \multirow{2}{*}{T$_2$ }  & Original & 64.76 & \textbf{68.47} \\
 & Challenging Illumination & 46.02 & \textbf{62.28} \\ \hline

\end{tabular}
\end{center}
\vspace{-0.25cm}
\end{table}

In unstructured environments, lighting conditions are uncontrollable and unknown in advance. Due to the movement of the target and the camera, varying illumination occurs during UAV monitoring. Our approach not only improves the overall performance of aerial object detection, but also improves the detection under varying illumination conditions. To prove this claim, we perform additional tests. We programmatically simulate difficult illumination conditions on images from test datasets. The quantitative results are presented in Table \ref{tab:illum}.

\begin{figure} [t]
    \centering
    \includegraphics[width=1.0\columnwidth]{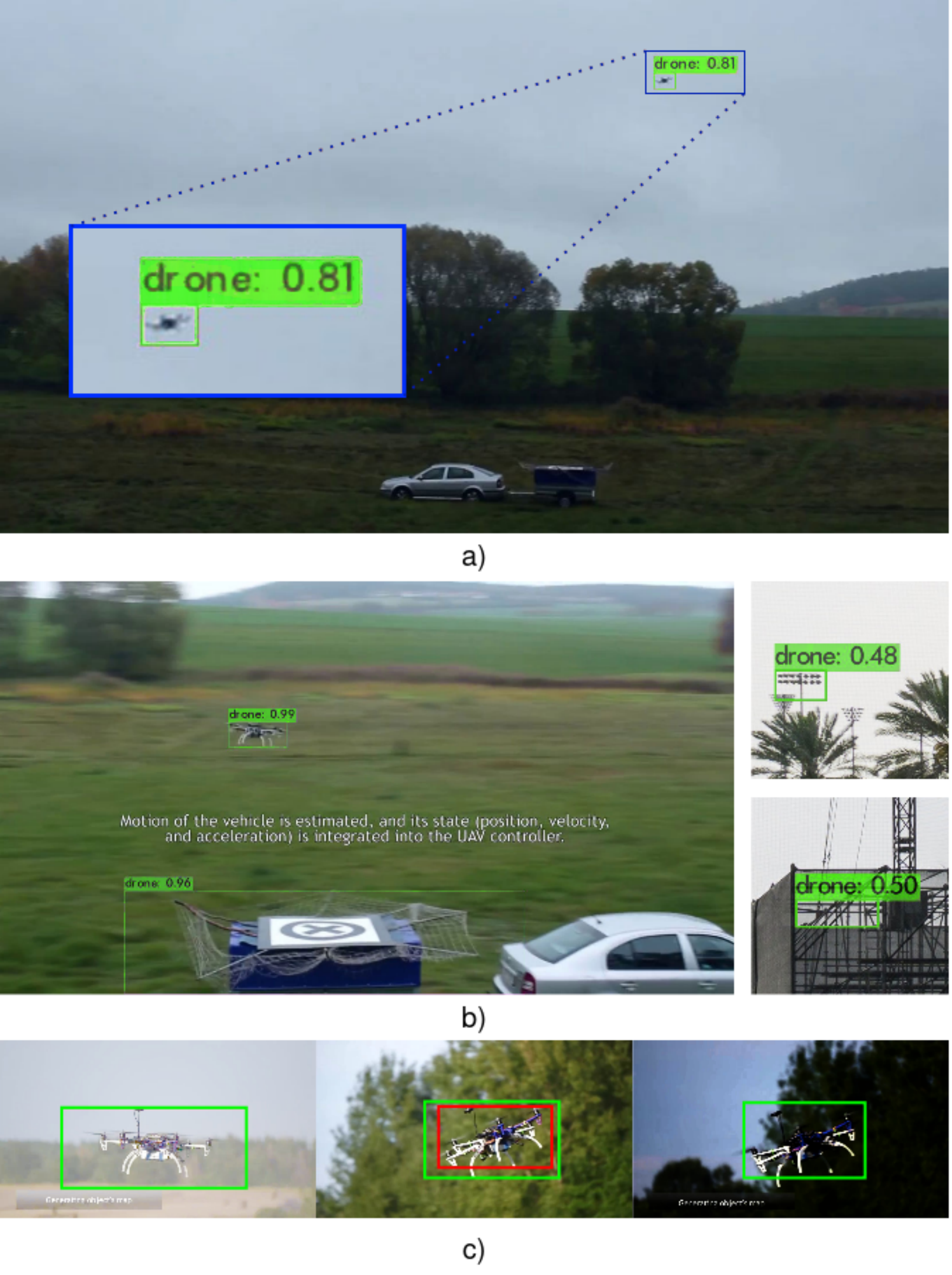}
    \caption{Detection results on images from the T$_2$ test dataset of the detector trained on purely synthetic data with unrealistic textures (green). (a) Example of successful detection of distant object where the detector trained purely on real images fails. (b) Example images of false positive detections caused by similarly shaped objects. (c) Detection results of the S-Eagle-B (red) and S-Eagle-T (green) detectors on a bright overexposed image, a normally illuminated image, and a dark underexposed image.}\vspace{-0.15cm}
    \label{fig:collage}
\end{figure}

On both datasets, the proposed method of texture-invariant object detectors improves the detection under changing illumination by 16 percentage points in mAP on T$_2$ dataset and 4 percentage points on UAV-Eagle dataset. The qualitative results can be seen in Fig. \ref{fig:collage}c. The synthetic texture-invariant detector (green bounding box) detects the target object under all conditions, in the bright overexposed image, in the normal illumination image, and in the underexposed dark image. The detector without our enhancement is only able to detect the object in a normally illuminated image. Improvements are also seen in the detection of distant objects, as can be seen in Fig. \ref{fig:collage}a, where the S-Eagle-T detector successfully detects the object, while the detector trained on R-UAV fails. The presented results show that the Sim2Air gap can be bridged with the proposed approach and that the accuracy of the real data can even be surpassed. On the other hand, since the detector trained on S-Eagle-T is more attentive to shape, there are situations in which similarly shaped objects lead the detector to false positive detection, as shown in Fig. \ref{fig:collage}b. This problem could be solved by including negative samples of shape-like objects in the training dataset. In general, more attention will be paid in the future to reduce false positive detections by using techniques such as flying distractors \cite{Tremblay2018}.

\section{Conclusion}
\label{sec:conclusion}

In this paper, a texture-invariant object representation for aerial object detection is presented and evaluated on several test datasets with real images. A procedural pipeline for generating synthetic datasets is developed, implementing the proposed technique of randomly assigning atypical textures to UAV models. The results of the evaluation of the synthetically generated datasets confirm that shape plays a greater role in aerial object detection. This is due to the imaging conditions of the aerial perspective in unstructured dynamic environments, where the texture of the object is difficult to discern. Quantitative and qualitative results show that the proposed approach outperforms baseline and real-world data in situations with difficult lighting and distant objects.


\addtolength{\textheight}{-0.1cm}   



\section*{ACKNOWLEDGMENT}
\small{This work has been supported by European Commission Horizon 2020 Programme through project under G. A. number 810321: Twinning coordination action for spreading excellence in Aerial Robotics - AeRoTwin. Work of Frano Petric was supported by the European Regional Development Fund under the grant KK.01.1.1.01.0009 (DATACROSS).}


\bibliographystyle{ieeetr}
\typeout{}
\balance
\bibliography{bibliography/bib}

\end{document}